\documentclass[twoside,11pt]{article}

\usepackage{blindtext}
\usepackage{marvosym}
\usepackage{booktabs}
\usepackage{pifont}

\makeatletter
\def\customsymbol#1{
    \ifcase\number\value{#1}
        \or *
        \or \Letter
    \else\@ctrerr
    \fi
}
\makeatother

\usepackage[preprint]{jmlr2e}

\usepackage{xcolor}
\usepackage{amssymb}
\usepackage{graphics}

\usepackage{lastpage}
\jmlrheading{23}{2022}{1-\pageref{LastPage}}{1/21; Revised 5/22}{9/22}{21-0000}{You \textit{et.al.}}

\ShortHeadings{\texttt{depyf}: Open the Opaque Box of PyTorch Compiler for Machine Learning Researchers}{You \textit{et.al.}}
\firstpageno{1}

\begin{document}

\hypersetup{
  colorlinks,
  citecolor=gray,
  filecolor=black,
  linkcolor=black,
  urlcolor=black
}


\title{\texttt{depyf}: Open the Opaque Box of PyTorch Compiler for Machine Learning Researchers}

\author{\name Kaichao You$^\dag$\thanks{This work is conducted during Kaichao You’s internship at Apple.} \email youkaichao@gmail.com \\
        \name Runsheng Bai$^\dag$ \email brs21@mails.tsinghua.edu.cn \\
        \name Meng Cao$^\S$ \email mengcao@apple.com \\
        \name Jianmin Wang$^\dag$ \email jimwang@tsinghua.edu.cn \\
        \name Ion Stoica$^\ddag$ \email istoica@cs.berkeley.edu \\
        \name Mingsheng Long$^\dag$\thanks{Mingsheng Long is the corresponding author.} \email mingsheng@tsinghua.edu.cn \\
$^\dag$ School of Software, BNRist, Tsinghua University, Beijing 100084, China\\
    $^\S$ AIML, Apple \quad $^\ddag$ Division of Computer Science, UC Berkeley, CA 94720-1776, USA\\ 
}

\editor{}

\maketitle

\begin{abstract}%
  PyTorch \texttt{2.x} introduces a compiler designed to accelerate deep learning programs. However, for machine learning researchers, adapting to the PyTorch compiler to full potential can be challenging. The compiler operates at the Python bytecode level, making it appear as an opaque box. To address this, we introduce \texttt{depyf}, a tool designed to demystify the inner workings of the PyTorch compiler. \texttt{depyf} decompiles bytecode generated by PyTorch back into equivalent source code, and establishes connections between in-memory code objects and their on-disk source code counterparts. This feature enables users to step through the source code line by line using debuggers, thus enhancing their understanding of the underlying processes. Notably, \texttt{depyf} is non-intrusive and user-friendly, primarily relying on two convenient context managers for its core functionality. The project is  \href{https://github.com/thuml/depyf}{\color[rgb]{0.5,0.5,0.5} openly available} and is recognized as a \href{https://pytorch.org/ecosystem/}{\color[rgb]{0.5,0.5,0.5} PyTorch ecosystem project}.
\end{abstract}


\vspace{-10pt}
\section{Introduction}

Deep learning has profoundly impacted our daily lives, especially with the recent advancements in Large Language Models (LLMs) like ChatGPT~\citep{schulman_chatgpt:_2022}. These models demand considerable computational resources, prompting the swift development of specialized hardware~\citep{lecun_deep_2019}, such as GPUs~\citep{markidis_nvidia_2018} and TPUs~\citep{jouppi_domain-specific_2020}. However, fully leveraging the capabilities of these advanced hardware is challenging. It requires in-depth knowledge of hardware-specific programming, exemplified by technologies like FlashAttention~\citep{dao_flashattention:_2022}. Such expertise often extends beyond the focus of machine learning researchers who concentrate on algorithm development. To bridge this gap, domain-specific deep learning compilers have been introduced~\citep{li_deep_2020}. These compilers are crafted to optimize deep learning programs for efficient operation on modern hardware. While these compilers simplify the optimization process, adapting them to maximize benefits remains a complex endeavor. This complexity highlights the ongoing tension between hardware advancements and software optimization in the rapidly evolving field of deep learning.

PyTorch~\citep{paszke_pytorch:_2019}, a widely-used deep learning framework among machine learning researchers, was traditionally imperative and user-friendly. To keep pace with recent hardware advancements and to enable better optimization for large-scale distributed training~\citep{rasley_deepspeed:_2020,shoeybi_megatron-lm:_2020}, PyTorch recently underwent a significant update, transitioning from PyTorch \texttt{1.x} to PyTorch \texttt{2.x}. This update included the integration of a built-in deep learning compiler, the \texttt{torch.compile}~\footnote{\url{https://pytorch.org/docs/stable/torch.compiler.html}} function. This addition narrows the gap for machine learning researchers in utilizing modern hardware, but a notable gap remains and is still challenging to bridge.

This paper first describes the challenges machine learning researchers face in understanding the PyTorch compiler, illustrated through a concrete example. It then discusses how the proposed tool addresses these challenges, concluding with practical usage examples and experimental results.

\vspace{-10pt}
\section{Challenges in Understanding the PyTorch Compiler}

\subsection{Dynamo: The Frontend of the PyTorch Compiler}

\begin{figure}
  \vspace{-10pt}
  \includegraphics[width=\textwidth]{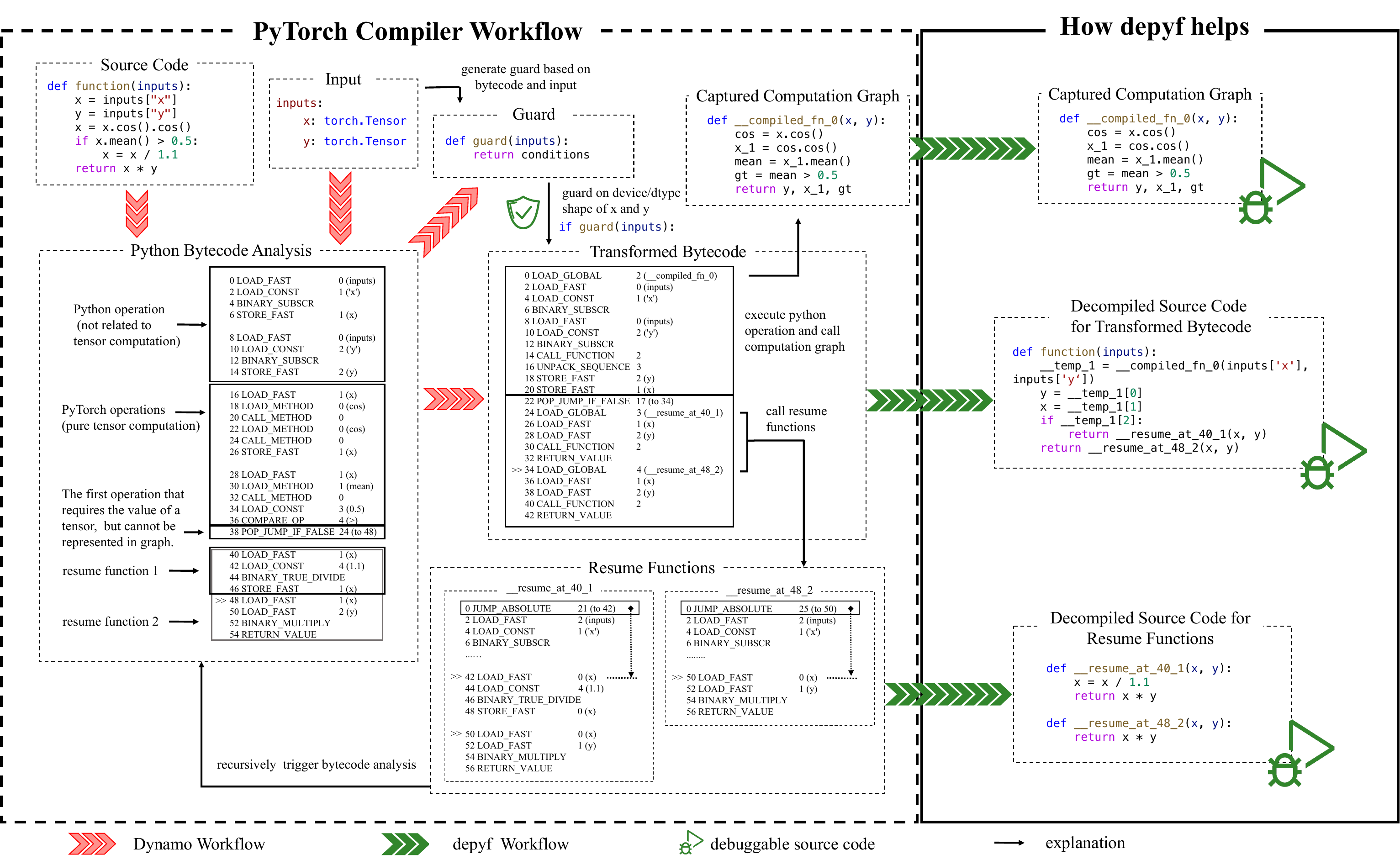}
  \vspace{-20pt}
  \caption{The workflow of the PyTorch compiler (left), and how \texttt{depyf} helps (right).}
  \label{fig:dynamo}
\end{figure}

The most complex component of the PyTorch compiler is its frontend named Dynamo. Dynamo's key functionality is to separate user code into distinct segments: pure Python code and pure PyTorch code, which forms the computation graph. Figure~\ref{fig:dynamo} (left) provides a detailed example of Dynamo's operation. This process involves three primary steps:
\begin{itemize}
  \setlength{\itemsep}{-5pt} 
  \item Identifying the first operation that cannot be represented in the computation graph but requires the value of a previously computed tensor in the graph. Examples include operations like \texttt{print}ing a tensor’s value or using a tensor’s value to determine the control flow in Python \texttt{if} statements.
  \item Dividing preceding operations into two segments: a computation graph focused solely on tensor computations and Python code dedicated to manipulating Python objects.
  \item Handling the subsequent operations as one or more new functions (referred to as \texttt{resume functions}) and recursively reinitiating the analysis described above.
\end{itemize}
Dynamo functions at the Python bytecode level (see \texttt{LOAD}, \texttt{JUMP}, \texttt{CALL} instructions in Figure~\ref{fig:dynamo}), which is a more fundamental level than Python source code. It's important to note that {\color{blue}very few machine learning researchers are proficient in interpreting this bytecode}.

\subsection{The Backend of the PyTorch Compiler}

After the frontend extracts a computation graph, the backend optimizes this graph and ultimately generates binary executables suitable for CPU, GPU, and TPU hardware. A computation graph in Python is a \emph{dynamically generated} function, meaning it must be \emph{executed in its entirety}. Consequently, users are {\color{blue}unable to employ debuggers for a step-by-step analysis of the function}. This becomes particularly challenging when the computation results in a \texttt{NaN} (Not a Number) error, as it precludes the possibility of tracing through the code line by line to identify the operation responsible for the numeric issue.

\section{Solution}


\textbf{Bytecode Decompilation:} The primary goal is to free machine learning researchers from the complexities of bytecode. The process of converting bytecode back into source code is called ``decompilation''. Before \texttt{depyf}, existing Python decompilers could transform Python bytecode into source code, but they have significant limitations:
\begin{itemize}
  \setlength{\itemsep}{-5pt} 
  \item They typically support only old versions of Python with limited compatibility.
  \item Designed for decompiling bytecode compiled from source code, they struggle with program-generated bytecode like that from PyTorch.
\end{itemize}
To overcome these issues, we created a new Python bytecode decompiler~\footnote{The name \texttt{depyf} stands for: \underline{de}compile \underline{Py}thon \underline{f}unctions. We focus on function bytecodes, which is also the main focus of the PyTorch compiler.} through symbolic execution of the bytecode. This approach requires handling only about two hundred types of Python bytecode, ensuring \emph{compatibility with all Python versions supported by PyTorch}.

Moreover, the core component of the PyTorch compiler, written in C, is replicated in Python within \texttt{depyf} to elaborate the underlying mechanisms for users.

\vspace{5pt}

\noindent \textbf{Function Execution Hijacking:} To facilitate line-by-line code execution with debuggers, the bytecode executed by Python must originate from an on-disk source code file. We utilize advanced Python features to intercept and replace critical function calls in PyTorch. This replacement involves dynamically generated functions with counterparts that include debugging information.

\vspace{5pt}

\noindent \textbf{Usage:} Using \texttt{depyf} is straightforward and non-intrusive. Users simply need to enclose their code within the context manager \texttt{with depyf.prepare\_debug()}. This action enables \texttt{depyf} to capture all internal PyTorch details in that context, including decompiled source code and the computation graph. For those wishing to step through decompiled code with debuggers, an additional context manager, \texttt{with depyf.debug()}, is available. Appendix~\ref{appendix:usage} gives more details about the usage.

\vspace{5pt}

\noindent \textbf{Overview:} Figure~\ref{fig:dynamo} provides an overview of the \texttt{depyf} process. More comprehensive details can be found on our documentation page~\footnote{\url{https://depyf.readthedocs.io/en/latest/}}. The advantages of \texttt{depyf} are threefold:
\begin{itemize}
  \setlength{\itemsep}{-5pt} 
  \item It offers a Python implementation analogous to PyTorch’s C implementation, aiding users in grasping the PyTorch compiler’s logic. (See \texttt{full\_code\_xxx.py} in Figure~\ref{fig:usage})
  \item It includes a Python bytecode decompiler that transforms bytecode into equivalent source code, helping users understand the transformed bytecode from PyTorch. (See \texttt{\_\_transformed\_xxx.py} in Figure~\ref{fig:usage})
  \item It hijacks critical functions in PyTorch, enabling users to step through computation graph functions line by line using debuggers. (See \texttt{\_\_compiled\_xxx.py} in Figure~\ref{fig:usage})
\end{itemize}

\begin{table}
  \centering
  \vspace{-20pt}
  \resizebox{\textwidth}{!}{
  \begin{tabular}{cccccc}
    \toprule
      Decompiler & Python 3.8 & Python 3.9 & Python 3.10 & Python 3.11 & PyTorch\\
      \midrule
      \href{https://github.com/rocky/python-decompile3}{decompyle3} & $90.6\%(77/85)$ & \ding{55} & \ding{55} & \ding{55} & \ding{55}\\
      \midrule
      \href{https://github.com/rocky/python-uncompyle6}{uncompyle6} & $91.8\%(78/85)$ & \ding{55} & \ding{55} & \ding{55} & \ding{55}\\
      \midrule
      \href{https://github.com/zrax/pycdc}{pycdc} & $74.1\%(63/85)$ & $74.1\%(63/85)$ & $74.1\%(63/85)$ & $67.1\%(57/85)$ & $19.3\%(27/140)$ \\
      \midrule
      \href{https://github.com/thuml/depyf}{depyf} & $100\%(85/85)$ & $100\%(85/85)$ & $100\%(85/85)$ & $100\%(85/85)$ & $100\%(140/140)$ \\
      \bottomrule
  \end{tabular}
  }
  \vspace{-10pt}
  \caption{Correctness of decompilers in Python and PyTorch tests.}
  \label{tab:decompiler_compare}
\end{table}

\vspace{-10pt}
\section{Experiments}

Table~\ref{tab:decompiler_compare} presents the compatibility status of various existing decompilers with Python and PyTorch. Detailed descriptions of these tests can be found in Appendices~\ref{appendix:model} and \ref{appendix:python}. Notably, {\texttt{depyf}} is \emph{the sole decompiler to successfully pass all the tests}. Our testing approach is conducted in a continuous integration manner, whereby every new commit undergoes testing against the nightly version of PyTorch across all supported Python versions. This proactive strategy allows us to identify and resolve any compatibility issues before the release of new PyTorch versions. Furthermore, we engage in discussions with the PyTorch team to propose solutions that maintain this compatibility.

Additionally, we have collected all the outputs from these experiments. This collection serves as a valuable resource for newcomers to PyTorch compilers, offering insights into the computational aspects of common deep learning models. More details can be found in Appendix~\ref{appendix:output}.

\vspace{-10pt}
\section{Conclusion}

In this paper, we introduced \texttt{depyf}, a novel tool designed to open the opaque box of the PyTorch compiler, facilitating machine learning researchers' understanding and adaptation to \texttt{torch.compile}.


\acks{
We thank the PyTorch team for helpful discussions and guidance on many internal details. Special thanks for Jason Ansel, Horace He, and Edward Yang.

We thank many colleagues from the Machine Learning Group in Tsinghua University (THUML) for providing helpful feedback on the draft and documentation.

Kaichao You is partly supported by
the Apple Scholar in AI/ML.

This work is supported by the National Natural Science Foundation of China through the Fund for Creative Research Groups (62021002) and Fund for Excellent Young Scholars (62022050).
}

\appendix

\section{Usage}
\label{appendix:usage}

We provide two convenient context managers for users: \texttt{with depyf.prepare\_debug()} and \texttt{with depyf.debug()}. The first one will capture all the calls to functions using \texttt{torch.compile}, and dump many internal details in a directory specified by users (\emph{i.e.}, the argument of \texttt{with depyf.prepare\_debug()}). The second one will pause the program for users to set breakpoints in the dumped source code, and any call to functions related with \texttt{torch.compile} can be stepped through line by line using standard Python debuggers.

There are three types of source code dumped by \texttt{depyf}: computation graphs (prefixed by \texttt{\_\_compiled}), decompiled source code (prefixed by \texttt{\_\_transformed}), and descriptive source code (prefixed by \texttt{full\_code}).

\begin{figure}[h]
  \includegraphics[width=\textwidth]{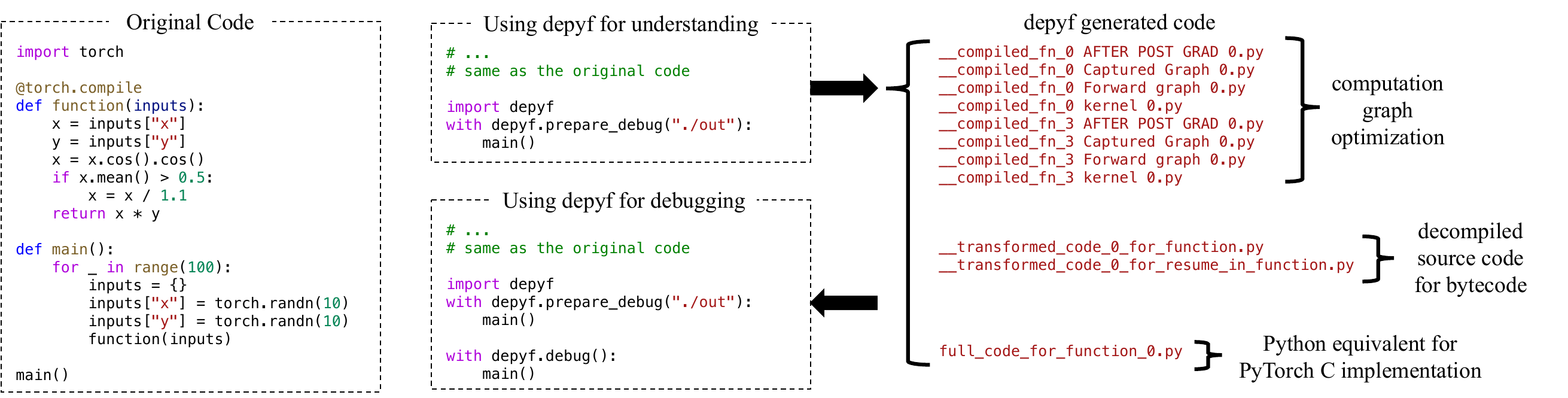}
  \caption{Two usage of \texttt{depyf}.}
  \label{fig:usage}
\end{figure}

\section{Tested PyTorch Models}
\label{appendix:model}

This section lists all of the PyTorch models we test in Table~\ref{tab:decompiler_compare}. These models come from three suites of deep learning models: TorchBench~\citep{Constable_TorchBench_A_collection_2020} collects models from famous (highly cited projects as ranked by \url{https://paperswithcode.com/}) machine learning repositories like Segment Anything~\citep{kirillov_segment_2023} and SuperSloMo~\citep{jiang_super_2018}; Huggingface Transformers~\citep{wolf_transformers:_2020} is the most popular library for transformers models including LLaMA~\citep{touvron_llama:_2023} and BERT~\citep{devlin_bert:_2019}; TIMM~\citep{wightman_pytorch_2023} is the most popular library for computer vision models including ResNet~\citep{he_deep_2016} and ViT~\citep{dosovitskiy_image_2021}.

To be specific, the models include:

\begin{itemize}
\setlength{\itemsep}{-5pt} 
  \item \texttt{BertForMaskedLM, BertForQuestionAnswering, BERT\_pytorch, hf\_Bert}~\citep{devlin_bert:_2019} 

  \item \texttt{AlbertForMaskedLM, AlbertForQuestionAnswering}~\citep{lan2020albert} 

  \item \texttt{AllenaiLongformerBase}~\citep{beltagy2020longformer}
  
  \item \texttt{BartForCausualLM, BartForConditionalGeneration, hf\_Bart}~\citep{lewis2019bart} 

  \item \texttt{BlenderbotForCausualLM, BlenderbotForConditionalGeneration, Blenderbot- SmallForCausualLM, BlenderbotSmallForConditionalGeneration}~\citep{shuster2022blenderbot} 

  \item \texttt{CamemBart}~\citep{Martin_2020} 

  \item \texttt{DebertaForMaskedLM, DebertaForQuestionAnswering, DebertaV2ForMaskedLM, \\ DebertaV2ForQuestionAnswering}~\citep{he2021deberta} 
  
  \item \texttt{DistilBertForMaskedLM, DistilBertForQuestionAnswering}~\citep{sanh2020distilbert} 

  \item \texttt{DistilGPT}~\citep{radford2019language, sanh2020distilbert}

  \item \texttt{ElectraForCausalLM, ElectraForQuestionAnswering}~\citep{clark2020electra} 

  \item \texttt{GPT2ForSequenceClassification, hf\_GPT2}~\citep{radford2019language} 

  \item \texttt{GPTJForCausalLM, GPTJForQuestionAnswering}~\citep{radford2022robust}

  \item \texttt{GPTNeoForCausalLM, GPTNeoForSequenceClassification}~\citep{gao2020pile}

  \item \texttt{LayoutLMForMaskedLM, LayoutLMForSequenceClassification}~\citep{Xu_2020} 
  
  \item \texttt{M2M100}~\citep{fan2020englishcentric} 

  \item \texttt{MBartForCausalLM, MBartForSequenceClassification}~\citep{liu2020multilingual} 

  \item \texttt{MT5ForConditionalGeneration}~\citep{xue2021mt5} 

  \item \texttt{MegatronBertForMaskedLM, MegatronBertForQuestionAnswering}~\citep{fan2020englishcentric} 

  \item \texttt{MobileBertForMaskedLM, MobileBertForQuestionAnswering}~\citep{sun2020mobilebert} 

  \item \texttt{OPTForCausalLM}~\citep{zhang2022opt} 

  \item \texttt{PLBartForCausalLM, PLBartForConditionalGeneration}~\citep{ahmad2021unified} 

  \item \texttt{PegasusForCausalLM, PegasusFOrConditionalGeneration}~\citep{zhang2020pegasus} 

  \item \texttt{RoBERTaForCausalLM, RoBERTaForQuestionAnswering}~\citep{liu2019roberta} 

  \item S2T2~\citep{lin2022semisupervised}  

  \item \texttt{T5ForConditionalGeneration, T5Small, hf\_T5}~\citep{raffel2023exploring} 

  \item \texttt{TrOCRForCausalLM}~\citep{li2022trocr} 

  \item \texttt{XGLMForCausalLM}~\citep{lin2022fewshot} 

  \item \texttt{XLNetLMHeadModel}~\citep{yang2020xlnet} 

  \item \texttt{YituTechConvBert}~\citep{jiang2021convbert} 

  \item \texttt{gluon\_inception\_v3, inception\_v3}~\citep{szegedy2015rethinking} 

  \item \texttt{adv\_inception\_v3}~\citep{kurakin2018adversarial}

  \item \texttt{beit\_base\_patch16\_224}~\citep{bao2022beit} 

  \item \texttt{botnet26t\_256}~\citep{srinivas2021bottleneck} 

  \item \texttt{eca\_botnext26ts\_256, sebotnet33ts\_256}~\citep{srinivas2021bottleneck, wightman2021resnet}

  \item \texttt{cait\_m36\_384}~\citep{touvron2021going} 

  \item \texttt{coat\_lite\_mini}~\citep{xu2021coscale} 

  \item \texttt{convit\_base}~\citep{d_Ascoli_2022} 

  \item \texttt{convmixer\_768\_32}~\citep{Ng_2022} 

  \item \texttt{convnext\_base}~\citep{liu2022convnet} 

  \item \texttt{crossvit\_9\_240}~\citep{chen2021crossvit} 

  \item \texttt{cspdarknet53}~\citep{wang2019cspnet, bochkovskiy2020yolov4} 

  \item \texttt{deit\_base\_distilled\_patch16\_224}~\citep{touvron2021training} 

  \item \texttt{dla102}~\citep{yu2019deep} 

  \item \texttt{dm\_nfnet\_f0, nfnet\_l0, timm\_nfnet}~\citep{brock2021highperformance}: 

  \item \texttt{dpn107}~\citep{chen2017dual} 

  \item \texttt{eca\_halonext26ts}~\citep{vaswani2021scaling, wightman2021resnet} 
  
  \item \texttt{ese\_vovnet19b\_dw, timm\_vovnet}~\citep{lee2020centermask} 

  \item \texttt{fbnetc\_100, fbnetv3\_b}~\citep{wu2019fbnet} 

  \item \texttt{gernet\_l}~\citep{lin2020neural} 

  \item \texttt{ghostnet\_100}~\citep{han2020ghostnet} 
  
  \item \texttt{mixer\_b16\_224, gmixer\_24\_224}~\citep{tolstikhin2021mlpmixer} 

  \item \texttt{gmlp\_s16\_224}~\citep{liu2021pay} 

  \item \texttt{hrnet\_w18}~\citep{wang2020deep} 

  \item \texttt{jx\_nest\_base}~\citep{zhang2021nested} 

  \item \texttt{lcnet\_050}~\citep{cui2021pplcnet}

  \item \texttt{levit\_128}~\citep{graham2021levit} 

  \item \texttt{mixnet\_l, tf\_mixnet\_l}~\citep{tan2019mixconv} 

  \item \texttt{mnasnet\_100, mnasnet1\_0}~\citep{tan2019mnasnet} 

  \item \texttt{mobilenetv2\_100, mobilenet\_v2}~\citep{sandler2019mobilenetv2, wightman2021resnet} 

  \item \texttt{mobilenetv3\_large\_100, mobilenet\_v3\_large}~\citep{howard2019searching}
  
  \item \texttt{mobilevit\_s}~\citep{mehta2022mobilevit} 

  \item \texttt{pit\_b\_224}~\citep{heo2021rethinking} 
  
  \item \texttt{pnasnet5large}~\citep{liu2018progressive} 

  \item \texttt{poolformer\_m36}~\citep{yu2022metaformer} 

  \item \texttt{regnety\_002, timm\_regnet}~\citep{radosavovic2020designing} 

  \item \texttt{repvgg\_a2}~\citep{ding2021repvgg} 

  \item \texttt{res2net101\_26w\_4s, res2net50\_14w\_8s, res2next50, resnet18, resnet50}~\citep{Gao_2021} 

  \item \texttt{resmlp\_12\_224}~\citep{touvron2021resmlp} 

  \item \texttt{resnest101e, timm\_resnest}~\citep{zhang2020resnest} 

  \item \texttt{rexnet\_100}~\citep{han2021rethinking} 

  \item \texttt{selecsls42b}~\citep{Mehta_2020} 

  \item \texttt{spnasnet\_100}~\citep{stamoulis2019singlepath} 

  \item \texttt{swin\_base\_patch4\_window7\_224}~\citep{liu2021swin} 

  \item \texttt{swsl\_resnext101\_32x16d, resnext50\_32x4d}~\citep{xie2017aggregated} 

  \item \texttt{tf\_efficientnet\_b0, timm\_efficientnet}~\citep{tan2020efficientnet, xie2020selftraining} 

  \item \texttt{tinynet\_a}~\citep{han2020model} 

  \item \texttt{tnt\_s\_patch16\_224}~\citep{han2021transformer} 

  \item \texttt{twins\_pcpvt\_base}~\citep{chu2021twins} 

  \item \texttt{visformer\_small}~\citep{chen2021visformer} 

  \item \texttt{vit\_base\_patch16\_224, timm\_vision\_transformer}~\citep{dosovitskiy2021image} 

  \item \texttt{volo\_d1\_224}~\citep{yuan2021volo} 

  \item \texttt{xcit\_large\_24\_p8\_224}~\citep{elnouby2021xcit} 

  \item \texttt{Background\_Matting}~\citep{lin2020realtime} 

  \item \texttt{LearningToPaint}~\citep{huang2019learning}

  \item \texttt{alexnet}~\citep{10.1145/3065386} 

  \item \texttt{dcgan}~\citep{radford2016unsupervised} 

  \item \texttt{densenet121}~\citep{huang2018densely} 

  \item \texttt{nvidia\_deeprecommender}~\citep{kuchaiev2017training} 

  \item \texttt{pytorch\_unet}~\citep{ronneberger2015unet} 

  \item \texttt{shufflenet\_v2\_x1\_0}~\citep{zhang2017shufflenet} 

  \item \texttt{squeezenet1\_1}~\citep{iandola2016squeezenet} 

  \item \texttt{vgg16}~\citep{simonyan2015deep}

\end{itemize}

\section{Tested Python Syntax}
\label{appendix:python}

We also collect commonly used Python features in the above models, and store them in a simple Python test with over 80 testcases in \url{https://github.com/thuml/depyf/blob/master/tests/test.py}.

\section{Collected Output}
\label{appendix:output}

We collect all the output from PyTorch in \url{https://github.com/thuml/learn\_torch.compile}. It includes many commonly used models, how PyTorch converts them, and what is the shape of tensors across training and inference. All details are in self-contained scripts.
\bibliography{sample}

\end{document}